\documentclass[sn-basic]{sn-jnl}

\usepackage{algorithm}
\usepackage{booktabs}
\usepackage{multirow}
\usepackage[normalem]{ulem}
\useunder{\uline}{\ul}{}

\jyear{2023}%

\theoremstyle{thmstyleone}%
\theoremstyle{thmstyletwo}%

\theoremstyle{thmstylethree}%

\begin{document}

\title[MSGNN]{MSGNN: Multi-scale Spatio-temporal Graph Neural
Network for Epidemic Forecasting}

\author[1]{\fnm{Mingjie} \sur{Qiu}}\email{1021010625@njupt.edu.cn}

\author[1]{\fnm{Zhiyi} \sur{Tan}}\email{tzy@njupt.edu.cn}

\author*[1]{\fnm{Bing-kun} \sur{Bao}}\email{bingkunbao@njupt.edu.cn}

\affil[1]{\orgdiv{The College of Telecommunications and Information Engineering}, \orgname{Nanjing University of Posts and Telecommunications}, \orgaddress{ \city{Nanjing}, \state{Jiangsu}, \country{China}}}


\abstract{Infectious disease forecasting has been a key focus and proved to be crucial in controlling epidemic. A recent trend is to develop forecasting models based on graph neural networks (GNNs). However, existing GNN-based methods suffer from two key limitations: (1) Current models broaden receptive fields by scaling the depth of GNNs, which is insufficient to preserve the semantics of long-range connectivity between distant but epidemic related areas. (2) Previous approaches model epidemics within single spatial scale, while ignoring the multi-scale epidemic patterns derived from different scales. \\ To address these deficiencies, we devise the \uline{M}ulti-scale \uline{S}patio-temporal \uline{G}raph \uline{N}eural \uline{N}etwork (MSGNN) based on an innovative multi-scale view. To be specific, in the proposed MSGNN model, we first devise a novel graph learning module, which directly captures long-range connectivity from trans-regional epidemic signals and integrates them into a multi-scale graph. Based on the learned multi-scale graph, we utilize a newly designed graph convolution module to exploit multi-scale epidemic patterns. This module allows us to facilitate multi-scale epidemic modeling by mining both scale-shared and scale-specific patterns. Experimental results on forecasting new cases of COVID-19 in United State demonstrate the superiority of our method over state-of-arts. Further analyses and visualization also show that MSGNN offers not only accurate, but also robust and interpretable forecasting result. }




\maketitle
\section{Introduction}\label{sec1}

The infectious diseases pose a serious hazard to global public health. For decades, epidemic modelers have been struggling in forecasting the spread of emerging infectious diseases, such as the Zika virus, the Ebola virus, and most recently, the COVID-19 virus. To put the virus in control, accurate forecasting of the epidemic is of great significance for both individuals and administrators. 

Recently, graph neural networks (GNNs) based methods have emerged as a promising way of combating epidemics. The core idea of these approaches is modeling the epidemic signals between different areas to learn the underlying patterns in historical data. For example, Earlier works \citep{a1,a4,a24} view geographic adjacency as the key factor of trans-area epidemic signals, then directly apply graph convolution on a static spatial graph. Some follow-on studies \citep{in1, in2, a3, in20} extend the spatial graphs to spatio-temporal graphs, where the trans-area signals are dynamically measured. Most recently, a technical trend is to incorporate more epidemic-related factors (e.g. the social network data) into trans-area epidemic signals. These factors enrich the existing spatio-temporal graph representations \citep{j1,a2,in21}. Benefiting from the learned epidemic evolution patterns, existing GNN-based methods achieve promising performance in epidemic forecasting task. 

\begin{figure}[!tbp]\label{fig1}
    \centering
    \includegraphics[width=0.95\linewidth]{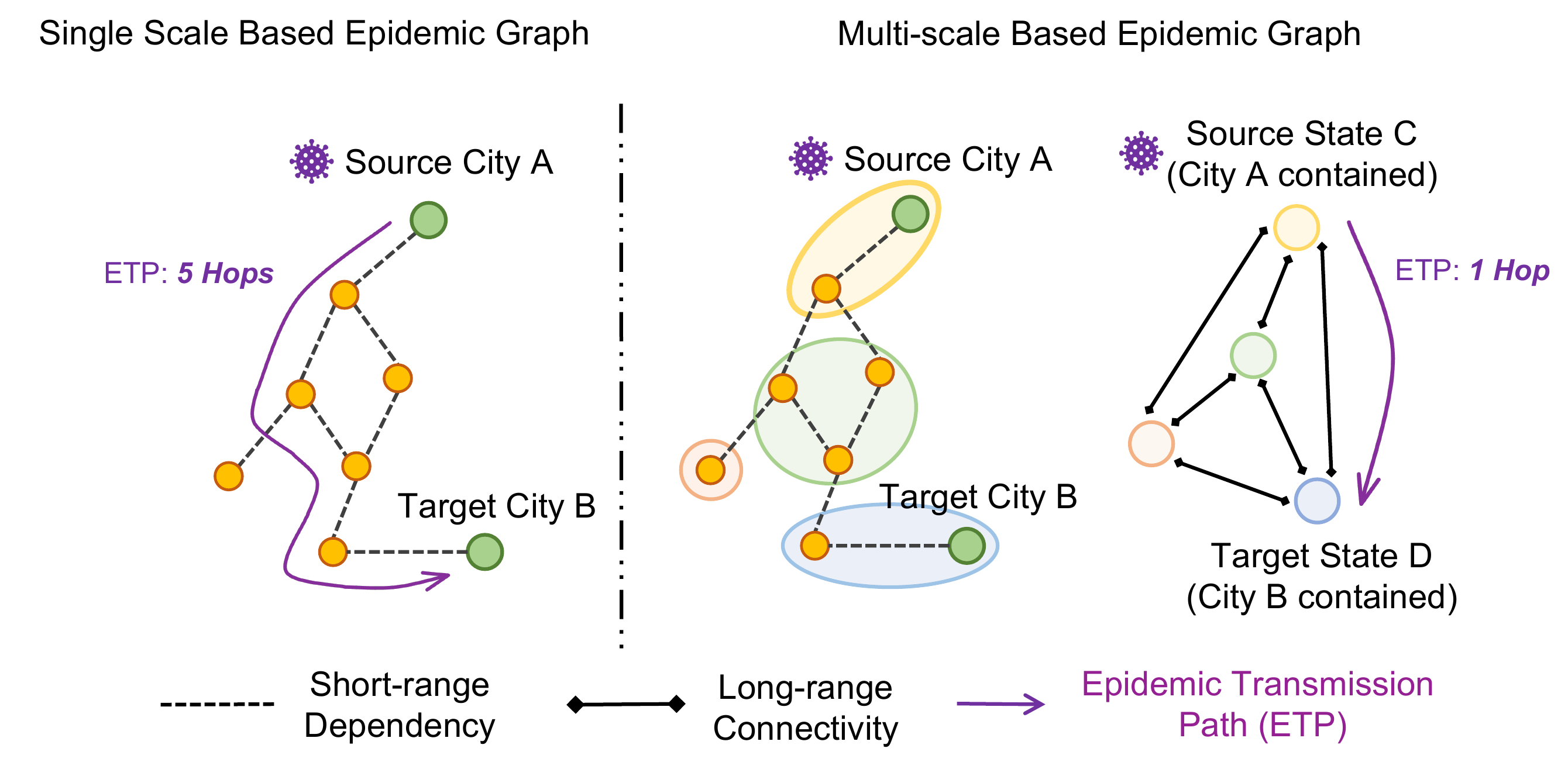}
    \caption{A Comparison between single-scale based and multi-scale based epidemic spatio-temporal graph. The additional macro scale helps model long-range connectivity, which significantly reduces the hops of epidemic transmission path between distant areas.}
\end{figure}

Despite their effectiveness, it is noticeable that most of the previous GNN-based methods model epidemics at single spatial scale. We suggest these single-scale based methods fall short in two respects: (1) \textbf{Failing to preserve long-range connectivity}. Existing works stack multiple convolution layers to aggregate information step by step, which dilutes the long-range connectivity between distant areas. As shown in the left side of Figure \ref{fig1}, in single scale view, the epidemic transmission path from city A to city B goes through five hops, which means the trans-area epidemic signal may mix up with noise from another four cities. More importantly, the receptive fields of existing GNN-based methods are significantly limited due to \textit{oversmoothing} issues \citep{in22}. These studies fail to preserve the semantics of long-range connectivity, i.e., the high-order relation between distant but epidemic related areas, leading to limited performance. (2) \textbf{Ignoring the multi-scale epidemic patterns}. Previous GNN-based works only exploit epidemic patterns within single spatial scale. However, an important fact has been ignored: the epidemic evolves simultaneously at different scales and reflects multiple epidemic evolution patterns. For example, the fine-grained epidemic evolution depicts local epidemic evolution pattern, while the coarse-grained evolution contains broader regional epidemic pattern. Such multi-scale patterns provide useful information that can aid accurate epidemic forecasting. Therefore, it is important to extract epidemic patterns from different scales and conduct multi-scale modeling. Unfortunately, none of these works take these multi-scale epidemic patterns into consideration.    

To this end, we propose the Multi-scale Spatio-temporal Graph Neural Network (MSGNN) for infectious disease forecasting. The model contains two components to solve the fore-mentioned challenges correspondingly: (1) \textbf{Graph Structure Learning Module}. Since the semantic of long-range connectivity is hard to preserve through single-scale modeling, in this module, we construct a multi-scale spatio-temporal graph to deal with different epidemic relations. As shown on the right side of Figure \ref{fig1}, the proposed multi-scale spatio-temporal graph contains two scales. In micro scales, we define the trans-area signal as short-range dependency, which is utilized to express fine-resolution spatial topology. Furthermore, considering the broader spatial effect of macro scale, we view trans-regional signals between states as long-range connectivity, which significantly reduce the epidemic transmission hops between distant areas. All signals are dynamically self-adjusted along with epidemic evolution, forming a multi-scale spatio-temporal graph as output. (2) \textbf{Multi-scale Graph Convolution Module}. Considering the discrepancy between different spatial scales, directly conducting inter-scale aggregation is not a viable option. Therefore, we devise a multi-scale information aggregation scheme to distill scale-specific and scale-shared parts from multi-scale patterns. This scheme first applies scale-specific message passing to obtain aggregated features at each scale. Then, the scheme distills the scale-shared epidemic patterns from multi-scale spatio-temporal graph. Finally, a multi-scale fusion block is introduced to integrate multi-scale features based on scale-shared patterns. These fused representations serve as the final output for forecasting. 

To verify the effectiveness of proposed model, we choose the latest and one of the most prevalent infectious disease, the COVID-19 virus as the validation epidemic. We conduct a comprehensive result evaluation on the COVID-19 dataset of the US, which contains two administrative levels, i.e. county level for micro scale and state level for macro scale. The result shows that our model outperforms state-of-art method in terms of both accuracy and robustness. We also organize another discussion part to explore the effect of three spatial dependencies. Our contributions can be summarized as follows: 

\begin{itemize}
    \item We propose an innovative multi-scale forecasting framework called MSGNN, which provides a novel epidemic modeling view that takes multiple administrative levels into account. 
    \item We design a multi-scale graph structure learning module to dynamically model epidemic relations at different scales, then further form a multi-scale spatio-temporal graph for epidemic forecasting.  
    \item We design a multi-scale graph convolution network powered by a novel multi-scale information fusion scheme. This scheme distinguishes the scale-shared and scale-specific patterns, then further encodes them into representation for forecasting. 
\end{itemize}

\section{Related Work}\label{sec2}
Many works have put their efforts into epidemic forecasting, which can be generally categorized into three diagrams: compartmental models, time series forecasting methods, and spatio-temporal forecasting methods.

The compartmental model is a classical modeling framework in epidemiology and widely adopted in infectious disease forecasting task \citep{in3,in4,a5,in5}. The main idea of the compartmental model is to divide the population into different compartments such as S (susceptible), I (infectious), and R (recovered), and then model the transition among these compartments with differential equations \citep{j2,a6}. In response to the COVID-19 outbreak, there emerges plenty of compartmental models specially designed according to COVID-19 virus characteristics. The main advantage of these models can be summarized to their outstanding interpretability \citep{in6, a7}, as they are able to produce interpretable coefficients such as infection rate, death rate, etc. However, existing compartmental models are limited to a pre-fixed basic reproduction rate, failing to learn the epidemic evolution patterns at different stages.

Time series forecasting methods reshape epidemic data into time series, turning the epidemic forecasting task into an auto-regressive problem. Common time series forecasting methods include classical statistical methods, hybrid methods and deep learning methods. The widely adopted classical statistical methods of include Auto Regressive Integrated Moving Average (ARIMA)\citep{a8,a9}, Multi-Linear Regression (MLR) \citep{a10}, etc. The hybrid methods are combinations of classical methods and other methods such as machine learning \citep{a11,a12} and boosting trees \citep{in7}. With the advancement of deep learning, some studies started to introduce neural networks such as the Long-Short Term Memory (LSTM) \citep{a13,a14,a15} network to model epidemic evolution. Time series forecasting methods are capable of handling complicated temporal dependencies. However, the future epidemic trend of a specific region is not only conditioned on the local spreading, but also influenced by transmissions from other epidemic-related areas. Existing time series forecasting methods ignore the potential spatial dependencies between epidemic-related areas, leading to limited performance. 

Recently, graph neural networks have attracted attention because of their strong capability in dealing with non-Euclidean space data \citep{in8, in9}. With the assistance of GNN, some studies extend the time series forecasting models with spatial interactions, forming a new paradigm called spatio-temporal forecasting \citep{in10,in11,in12,in13}. This paradigm are also applied in traffic forecasting \citep{a16,a17,in14} and weather forecasting \citep{in15}. Also, existing studies try to leverage spatio-temporal forecasting in epidemic forecasting \citep{a1,in16,book1}. Based on time series, these methods are capable of modeling spatial dependencies. However, due to the oversmoothing issue, their aggregation fields are limited within two hops, which means the ignorance of long-range connectivity. Moreover, the influence of multi-scale epidemic patterns are also neglected in existing epidemic modeling works. 

To the best of our knowledge, our work is the first spatio-temporal forecasting study that founded on multi-scale epidemic modeling. Based on the multi-scale modeling view, we are capable of handling volatile long-range connectivity and multi-scale epidemic evolution patterns.

\section{Preliminary}\label{sec3}

\textbf{Epidemic signals.} The epidemic development of a specific region is not only determined by local disease spreading, but also influenced by other related areas. We summarize two kinds of epidemic signals. Firstly, the short-range dependency happens between nearby areas, denoted by trans-area epidemic signal matrix $A_{c}$; secondly, the underlying long-range connectivity, denoted by trans-regional epidemic signal matrix $A_{s}$. 

\noindent
\textbf{Multi-scale spatio-temporal graph.} Multi-scale spatio-temporal graph is a hierarchical graph containing multiple scales, denoted by $G=(V,E)$. The node set $V$ can be further decomposed into $V=V_{c} \cup V_{s}$, where $V_c$ and $V_s$ respectively represent node set of micro and macro scale. Similarly, the edge set $E$ can also be divided into $E=E_{s} \cup E_{c}$ to represent different edges at multiple scales. Notice both $V$ and $E$ are time varying, which is also an important feature of spatio-temporal graph. 

\noindent
\textbf{Infectious Disease Forecasting.} Given a region $i$ and time step $t$, we define the input series as $X_{i}(t)\in \mathbb{R}^{L_{b}\times C}$ as $X_{i}(t)=(x_{i,t-L_{b}+1},x_{i,t-L_{b}+2},...,x_{i,t})$, where $L_{b}$ is the number of look-back time steps, $C$ represents the feature dimension of input series, $x_{i,t}\in \mathbb{R}^{C}$ is the daily feature at time step $t$. The output series is similar to input series, we denote it as $\hat{Y}_{i}(t)\in \mathbb{R}^{L_{a} \times D}$, where $\hat{Y}_{i}(t)=(\hat{y}_{i,t+1},\hat{y}_{i,t+2},...,\hat{y}_{i,t+L_{a}})$, and $L_{a}$ is the number of look-ahead time steps, $D$ is the output feature dimension, $\hat{y}_{i,t+1}$ is the output value of new cases at time step $t+1$. For simplicity, we omit the time step identifier $t$ in the rest of paper.

\begin{figure}[!t]
    \centering
    \includegraphics[width=1\linewidth]{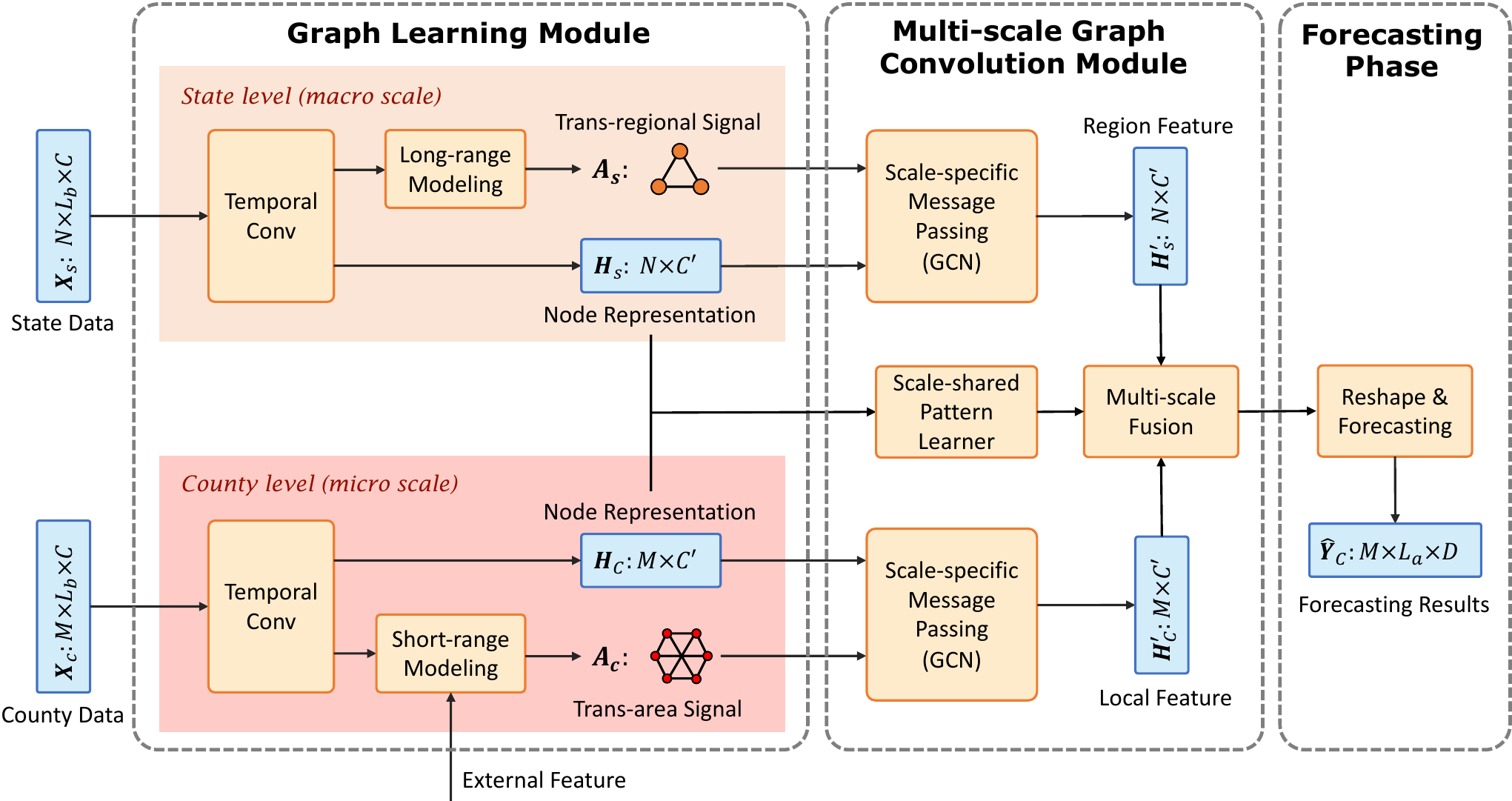}
    \caption{Pipeline of the proposed MSGNN. The pipeline consists of two main components, i.e. the graph learning module and the multi-scale graph convolution module.}
    \label{fig2}
\end{figure}

\section{Methods}\label{sec4}

As demonstrated in Figure \ref{fig2}, the proposed method is a multi-scale epidemic modeling process, which contains two main components. The pipeline starts with the graph learning module, which takes the state and county level epidemic data as input. The module first conducts temporal convolution to obtain node representation, then long and short range modeling blocks are applied to produce the epidemic signals for both macro and micro scales. Based on the epidemic signals and node representations, we construct a multi-scale spatio-temporal graph. The next part is the multi-scale graph convolution module, which adopts a new multi-scale aggregation scheme. It first applies message passing to embedding scale-specific patterns, forming region and local features. Afterwards, a learner is developed to distill scale-shared patterns through graph node representations. Afterwards, the region and local features are further integrated by a newly designed multi-scale fusion block, which takes these patterns into consideration. Finally, a forecast phase is applied to generate forecasting results. 

\subsection{Graph Learning Module}
The adaptive graph generation module is implemented by three components, including temporal convolution blocks, long-range modeling block and short-range modeling block. 

\subsubsection{Temporal Convolution Block}
The temporal convolution module is designed to extract local temporal features from historical epidemic time series.

For each scale, our goal is to encode the input time series $X$ into an encoded vector $H$. Theoretically, the temporal convolution block can be replaced by any recurrent structures such as RNN, GRU, or LSTM. However, limited by fixed kernel size, these naive approaches fail to dynamically encode volatile epidemic dynamics. To this end, we customize a more flexible temporal convolution block based on N-Beats \citep{in17} to reach our best practice. Differ from the naive N-Beats which only allows single time series as input, we additionally introduce more epidemic related series and external features, as shown in Equation \ref{eq1}. 

\begin{equation}
\label{eq1}
  \begin{aligned}
    X_{s} &= \operatorname{FC}\left(x_{t-l_{b}:t}^{s}, d_{t-l_{b}:t}\right) \oplus \mathcal{I}_{s} \in \mathbb{R}^{N \times L_{b} \times C} , \\
    X_{c} &= \operatorname{FC}\left(x_{t-l_{b}:t}^{c}, d_{t-l_{b}:t}\right) \oplus \mathcal{I}_{c} \in \mathbb{R}^{M \times L_{b} \times C};
  \end{aligned}
\end{equation}

\noindent
where $x_{t-l_{b}:t}$ is the input epidemic time series, $d_{t-l_{b}:t}$ is another time series including weekdays, holidays and date information, $\mathcal{I}$ is the location identity embedding for each county or state. We first utilize a fully connected layer to integrate date features into original time series, then an augmentation operation is applied to introduce the location identity embedding. This customized procedure allows the block to learn epidemic patterns produced by specified time and locations, which helps boosting the performance. Next, we apply the temporal convolution scheme as follows. 

\begin{equation}
\label{eq2}
  \begin{aligned}
    H_s &= \operatorname{MaxPool}\left[\operatorname{TC}(X_{s})\right] \in \mathbb{R}^{N \times C'},  \\
    H_c &= \operatorname{MaxPool}\left[\operatorname{TC}(X_{c})\right] \in \mathbb{R}^{M \times C'};
  \end{aligned}
\end{equation}

\noindent
where $TC$ is a temporal encoder based on N-Beats. We further conduct a max-pooling operation on the backcast output of temporal encoder to produce $H_{l}$ as encoded feature. 

\subsubsection{Long-Range Modeling Block}
The long-range modeling block is developed to directly capture the long-range connectivity between distant areas. This block takes the output of macro scale temporal convolution block $H_s$ as input, and produces an adjacency matrix $A_s$ representing the connectivity between different states.  

For specified states $l,k \in V_{s}$, we denote the pairwise connectivity as $[A]^{l,k}_{s}$. To dynamically measure the volatile long-range connectivity, we choose the real-time node representation $H_{l}, H_{k}$ as the key factor, which can be expressed as Equation \ref{eq3}. 

\begin{equation}
\label{eq3}
  \begin{aligned}
    \relax [A]^{l,k}_{s} &= f\left[\theta_{s}^{T} \cdot \operatorname{Concat}(H_{l},H_{k})\right]; \\
  \end{aligned}
\end{equation}

\noindent
where $\theta_{s}^{T} \in \mathbb{R}^{2 \times C'}$ is trainable mapping vectors, $f$ is the activation function. The idea of long-range connectivity modeling is mapping concatenated features into an adjacency edge scalar value. Due to the evolving node representations, the connectivities are also self-adjusted according to the real-time epidemic situation. In this block, the obtained long-range connectivity is determined by epidemic relations among different states. For those distant but epidemic-related areas, these newly introduced long-range connectivities can create a directly connected graph edge, thus breaking the spatial limitations. 

\subsubsection{Short-Range Modeling Block}
Aside from the long-range connectivity, we also elaborate on another trans-regional signal called short-range dependency. The short-range dependency are estimated at micro scale, which depicts the fine-resolution geographic topology for county level. Moreover, because of finer resolution, the short-range dependencies may be affected by other factors, such as geographic distance, administration boundary, etc. 

For counties $i,l$ in micro scale, we denote the short-range dependency as $[A]^{i,l}_{c}$. Differ from the long-range connectivity, we introduce extra signals to assist constructing the adjacency matrix, which can be expressed as Equation \ref{eq4}. 

\begin{equation}
\label{eq4}
  \begin{aligned}
    \relax [A]^{i,j}_{c} &= f\left[\theta_{c}^{T} \cdot \operatorname{Concat}(H_{i}, H_{j}, \mathcal{I}_{i}, \mathcal{I}_{j} )\right]+\frac{1}{\sqrt{\delta_{i,j}}};
  \end{aligned}
\end{equation}

\noindent
where $\theta_{c}^{T} \in \mathbb{R}^{4 \times C'}$ is a trainable mapping vector, $f$ is the activation function, $\mathcal{I}$ is the identity embedding for each location, and $\delta_{i,j}$ is the geographic distance between county $i$ and $j$. Apart from measuring epidemic relation, we additionally add the identity information to introduce the administration boundary information, which means counties in the same state will create stronger short-range dependency. Moreover, since the micro scale features narrower spatial effects, we add the geographic distance factor $\lambda$ to express the spatial limitations. Finally, to avoid noisy and redundant graph structure, we utilize $ReLU$ as the activation function to completely shutdown the adjacency edge between irrelevant counties. 

\subsection{Multi-scale Graph Convolution Module}
After the graph learning module, we obtain a new multi-scale spatio-temporal graph where each node represents regions and edges represent the epidemic signal. To model the epidemic evolution, a natural idea is to use graph convolution network to aggregate information. To this end, we propose the multi-scale graph convolution module. This module is composed by scale-specific message passing blocks, scale-shared patterns learner and a multi-scale fusion block, which forms a novel multi-scale information fusion scheme. In this module, the learned multi-scale spatio-temporal graph serves as the input, and the module outputs an epidemic representation for final forecasting. 

\subsubsection{Scale-specific Message Passing Block}
To deal with the scale-specific epidemic patterns, we implement the messaging passing block to deal with the epidemic information within single scale. The block takes the node representation $H$ and adjacency matrix $A$ as input. However, it is worth noting that $A$ is a diagonal-dominant matrix, which means the self-loops are much stronger than other connections. Since self-loops may do harm to the information aggregation between adjacent nodes \citep{in23}, we obtain new adjacency matrix $\tilde{A}_{s} \in \mathbb{R}^{N \times N}$ and $\tilde{A}_{c} \in \mathbb{R}^{M \times M}$ by zeroing the diagonal and performing normalization, as shown in Equation \ref{eq5}:  

\begin{equation}
\label{eq5}
\hat{A}=A-\operatorname{diag}\left({A}\right) , \quad \tilde{{A}}=\hat{{D}}^{-\frac{1}{2}} \hat{{A}} \hat{{D}}^{-\frac{1}{2}};
\end{equation}

\noindent
where $\hat{D}$ is the degree matrix of adjacency matrix $\hat{A}$. Then, we apply graph convolution at micro scale and macro scale respectively, so as to generate the local feature $H_{c}^{'} \in \mathbb{R}^{M \times C'}$ and region feature $H_{s}^{'} \in \mathbb{R}^{N \times C'}$ as Equation \ref{eq6}: 

\begin{equation}
\label{eq6}
\begin{aligned}
    H'_{c} &= \operatorname{GCN}(\tilde{A}_{c}, H_{c}) = \tilde{A}_{c}f\left( \tilde{A}_{c}H_{c}U_{1}\right)U_{2} \in \mathbb{R}^{M \times C'}; \\
    H'_{s} &= \operatorname{GCN}(\tilde{A}_{s}, H_{s}) = \tilde{A}_{s}f\left( \tilde{A}_{s}H_{s}U_{3}\right)U_{4} \in \mathbb{R}^{N \times C'};
\end{aligned}
\end{equation}

\noindent
where $U_{1}, U_{2}, U_{3}, U_{4} \in \mathbb{R}^{C' \times C'}$ are trainable weight matrix, $f$ is the activation function. Here we apply a two-layer GCN for both micro scale and macro scale. Note that the aggregation weight matrix for micro and macro scales are different, allowing the model to learn different epidemic patterns specified to certain scales. Through this way, the epidemic information are passed through multi-scale graph, which generates local features at micro scale and region features at macro scale. 

\subsubsection{Scale-shared Pattern Learner}
In this block, we utilize the obtained node representation to measure the scale-shared epidemic evolution pattern between state and county levels. Here we first construct a transfer matrix to identify the administrative affiliation between counties and their belonging states,  denoted as $Tran \in \mathbb{R}^{M \times N}$ as Equation \ref{eq7}: 

\begin{equation}
\label{eq7}
\left[Tran\right]_{i,l}=\left\{\begin{array}{l}
\frac{1}{\mathcal{N}(l)}, \text{ if county } i \text { affiliated to state } l\text{,} \\
0, \text { else;}
\end{array}\right.
\end{equation}

\noindent
where $\mathcal{N}(l)$ is the number of counties that belongs to state $l$. Next, we obtain the multi-scale epidemic patterns from different scales. For macro scale, we directly use the node representation as the epidemic patterns. For micro scale, We aggregate the node representations of counties that belong to the same state, as shown in the Equation \ref{eq8}: 

\begin{equation}
\label{eq8}
\begin{aligned}
    H^{Tran}_{c} &= (Tran)(H_{c}) \in \mathbb{R}^{N \times C'} 
\end{aligned}
\end{equation}

\noindent
According to the transfer matrix $Tran$, the produced $H^{Tran}_{c}$ is an averaged representation from all counties, e.g. the $l$-th row of $H^{Tran}_{c}$ refers to the averaged representation of all counties from state $l$. 

Since transfer representation $H^{Tran}_{c}$ and state representations $H_s$ are collected at different scales, they have different epidemic evolution patterns. To mine the latent scale-shared patterns, we use the Temporal Attention \citep{in24} to capture the cross-scale temporal correlations, as shown in Equation \ref{eq9}:

\begin{equation}
    \label{eq9}
    E = \left((H_s)^{T}U_{5}\right)U_{6}\left( H_{c}^{Tran}U_{7} + \beta \right)^{T} \in \mathbb{R}^{N \times N}
\end{equation}

\noindent
where $U_{5}, U_{7} \in \mathbb{R}^{N \times C'}$ and $U_{6} \in \mathbb{R}^{N \times N}$ are trainable weight matrix. Because of the dynamic node representations in spatio-temporal graph, the correlation matrix $E$ also changes along with epidemic evolution. 

\subsubsection{Multi-scale Fusion Block}
To incorporate local and region representations for final forecasting, we develop the multi-scale fusion block to comprehensively consider scale-shared and scale-specific patterns. We first devise a scheme to extract scale-shared epidemic representations from macro scale: 

\begin{equation}
    \label{eq10}
    \begin{aligned}
        \left[E'\right]_{i,j} &= \frac{\exp \left(\left[E\right]_{i, j}\right)}{\sum_{j\in(0,N)}^{i \in (0,N)} \exp \left(\left[E\right]_{i, j}\right)}, \\
        Att(H^{'}_{s}) &= E'H^{'}_{s} 
    \end{aligned}
\end{equation}

\noindent
The scale-shared pattern is extracted by correlation matrix $E$. Moreover, we further connect the scale-specific pattern $H'_c$ to the scale-shared part by Equation \ref{eq11}: 

\begin{equation}\label{eq11}
\begin{aligned}
H_{out} = \operatorname{Concat}\left[ Att(H'_{s}), H'_{c}\right] \in \mathbb{R}^{2C'}
\end{aligned}
\end{equation}

\subsection{Forecasting Module}
The forecasting module is implemented to forecast based on the output of multi-scale graph convolution network. 

To generate final forecasting results, we use a weight matrix $\theta_{f} \in \mathbb{R}^{L_{a} \times 2C'}$ to produce the final forecast $\hat{Y}_{i}$ of county $i$ with $L_a$ time steps ahead. During the training, we use the mean squared error of $\hat{Y}$ and $Y$ to form the loss function. For ground truth values $Y = \left\{y_{\left(t+1\right)}, \ldots, y_{\left(t+L_{a}\right)}\right\}$

\begin{equation}
\begin{aligned}
    \hat{Y} &= \theta_{f} \cdot  H_{out} ,\\
    loss &= MAE(\hat{Y}, Y) \\
    & =\frac{\sum_{i=1}^{L_{a}} \sum_{j=1}^N\lvert\left(\hat{y}_{\left(t+i\right)}^j-y_{\left(t+i\right)}^j\right)\rvert}{L_{a} * N}; 
\end{aligned}
\end{equation}

At the end of the method section, we summarize the algorithm of proposed MSGNN as shown in Algorithm \ref{algo1}.

\begin{algorithm}[!ht]
\caption{The MSGNN model for epidemic forecasting.}\label{algo1}
\begin{algorithmic}[1]
\Require The historical epidemic time series for micro scale (e.g. the county level in the US): ${X}_{c}=\left\{x^{c}_{t-L_{b}+1}, x^{c}_{t-L_{b}+2},\dots, x^{c}_{t}\right\} \in \mathbb{R}^{M \times L_{b} \times C}$;
\Require The historical epidemic time series for macro scale (e.g. the state level in the US): ${X}_{s}=\left\{x^{s}_{t-L_{b}+1}, \dots, x^{s}_{t}\right\} \in \mathbb{R}^{N \times L_{b} \times C}$;\\
Get the node representations of micro scale $H_{c} \in \mathbb{R}^{M \times C'}$ from $X_{c}$ by temporal convolution blocks;  \\
Get the node representations of macro scale $H_{s} \in \mathbb{R}^{N \times C'}$ from $X_{s}$ by temporal convolution blocks;  \\
Get trans-area epidemic signal matrix $A_{c} \in \mathbb{R}^{M \times M}$ from temporal feature $H_{c}$; \\
Get trans-regional epidemic signal matrix $A_{s} \in \mathbb{R}^{N \times N}$ from temporal feature $H_{s}$; \\
Get region feature $H'_s$ by applying message passing on graph $(A_{s}, H_{s})$; \\
Get local feature $H'_c$ by applying message passing on graph $(A_{c}, H_{c})$; \\
Get the combined feature $H_{out}$ from $H'_{c},H'_{s}, E$ by multi-scale fusion block; \\
Get forecasting \textit{Output} of MSGNN based on $H_{out}$; \\
\textbf{return} \textit{Output}; \\
Calculate the loss of MSGNN. 

\end{algorithmic}
\end{algorithm}

\section{Experiment}\label{sec5}

To validate the infectious disease modeling ability, we evaluate the proposed model by forecasting the COVID-19 epidemic, the latest and one of the most prevalent infectious diseases for decades. We elaborate the experiments mainly on the COVID-19 Forecast Hub \footnote{https://covid19forecasthub.org} (hereon referred as \emph{The Hub}), an official global challenge raised by United State Centers for Disease Control and Prevention (CDC) to accommodate weekly epidemic forecasting results from international groups. Since \emph{The Hub} released the forecasting results of all submitted models, we can easily get access to the performance of competitors and make comparison.

\subsection{Experimental Settings}
\subsubsection{Data} 
For COVID-19 epidemic data, we leverage daily reported cases and deaths from the Johns Hopkins University Center for Systems Science and Engineering (JHU CSSE) as the gold-standard data \footnote{https://github.com/CSSEGISandData/COVID-19}. The dataset collects incident cases and deaths from both state and county level in the United State. Following the location set defined by \emph{The Hub}, our experiment is conducted on 50 states and 3142 subordinated counties. The dataset date range is set from March 1, 2020 to July 1, 2021. The detailed statics are shown in Table \ref{tab1}. 

\begin{table}[!ht]
\centering
\caption{Detailed statics of the JHU CSSE dataset. }\label{tab1}
\begin{tabular}{@{}lcccccc@{}}
\toprule
                                                & \multicolumn{1}{l}{}           & \# location & \# dates & Min & Max   & Ave  \\ \midrule
\multicolumn{1}{l|}{\multirow{2}{*}{US-State}}  & \multicolumn{1}{c|}{confirmed} & 50          & 487      & 0   & 73854 & 1660 \\
\multicolumn{1}{l|}{}                           & \multicolumn{1}{c|}{deaths}    & 50          & 487      & 0   & 4417  & 450  \\ \midrule
\multicolumn{1}{l|}{\multirow{2}{*}{US-County}} & \multicolumn{1}{c|}{confirmed} & 3142        & 487      & 0   & 34497 & 62   \\
\multicolumn{1}{l|}{}                           & \multicolumn{1}{c|}{deaths}    & 3142        & 487      & 0   & 761   & 15   \\ \bottomrule
\end{tabular}
\end{table}

For geographical data, we directly use the geographic information in the JHU CSSE dataset, which contains the latitude and longitude of all regions in the US. The geographic data provides the state adjacency and county adjacency relationship information in the United State. 

For population data, we use the state and county population information (2019) in the United State to perform normalization to both confirmed and deaths data. 

\subsubsection{Evaluation Metrics}
As to evaluation metrics, we keep accordance with the requirements on \emph{The Hub}. Since the daily reported cases may be rather vibrating and unstable, we leverage weekly reported confirmed cases as our primary forecasting target. Our evaluation range is from February 2021 to July 2021. On each Sunday, we generate a forecasting result including one week, two weeks and three weeks ahead. In the five months evaluation period, we provides tens of forecasting results and calculating the performance indicators using three metrics, i.e. mean average error (MAE),  mean average percent error (MAPE) and root mean square error (RMSE), which can be calculated as Equation \ref{eq13}: 

\begin{equation}\label{eq13}
    \begin{aligned}
        MAE &=\frac{1}{n} \sum_{i=1}^{n}\lvert \hat{y}_i-y_i\rvert \text {, } \\
        MAPE &=\frac{100\%}{n} \sum_{i=1}^n\lvert \frac{\hat{y}_i-y_i}{y_i} \rvert \text{,} \\
        RMSE &=\sqrt{\frac{1}{n} \sum_{i=1}^n\left(\hat{y_i}-y_{i}\right)^2} \text{;}
    \end{aligned}
\end{equation}

\noindent
where $\hat{y}$ is the forecasting output, $y$ is the ground truth value, $n$ is the location number. As shown in Equation \ref{eq13}, the MAE measures the absolute difference between two values, averaged by the location number. The MAPE is also an error measure of two values in percentage terms. The RMSE is another frequently used metric, it represents the square root of the differences between predicted values and observed values. All the three metrics range in $[0,+\infty]$ and smaller values are better. 

For the whole evaluation period with over ten weeks, we conduct two experiments. We first average these per-week metrics to produce the main evaluation results containing MAE, MAPE, and RMSE. Moreover, we design another experiment called per-week evaluation to help understand the statistical distribution of forecasting results, so as to test the robustness of models. Besides, to keep consistency with the evaluation standard on \emph{The Hub}\footnote{https://covid19forecasthub.org/eval-reports/\#Incident\_Case\_Forecasts\_(county)}, we calculate the three metrics on the most 500 populous counties in the US (@500), i.e. $n=500$ in Equation \ref{eq13}. Furthermore, we utilize another relatively small dataset consisting of the most 100 populous counties (@100), i.e. $n=100$ in Equation \ref{eq13} to comprehensively evaluate the performances of all models. 

\subsection{Baselines}
To serve as the baselines, we choose a broad range of competitive COVID-19 forecasting models reported on \emph{The Hub}. It is worth noting that most of the models on \emph{The Hub} do not have their codes released, so instead of reproducing their forecasting results, we directly take their submitting result on the website. Furthermore, considering the frequency of submission and target forecasting location sets are varied, we set the filtering criteria as follows: candidate models should produce complete weekly outputs ranging from May, 2020 to July, 2021, also, the model must provide epidemic forecasting output of more than 3000 counties for full evaluation. We select 10 models satisfying the requirements as follows: 

\begin{itemize}
    \item \textbf{Microsoft-DeepSTIA} \citep{in2}: A deep spatio-temporal network, proposed by Microsoft Research. 
    \item \textbf{USC-SI\_kJalpha} \citep{a18}: A SIR model with vaccines and multiple variants, proposed by University of South California.
    \item \textbf{UVA-Ensemble} \citep{in18}: An ensemble model containing auto-regressive approach, SEIR approach and machine learning approach, proposed by University of Virginia. 
    \item \textbf{Google\_Harvard-CPF} \citep{in6}: A SEIR model fitted by machine learning approach, proposed by Google Cloud AI.
    \item \textbf{CEID\_Walk}\footnote{https://github.com/e3bo/random-walks}: A random walk model without drift, proposed by University of Georgia.
    \item \textbf{IowaStateLW-STEM} \citep{a19}: A quasi-likelihood approach via the penalized spline approximation, proposed by Iowa State University. 
    \item \textbf{JHUAPL-Bucky} \citep{a20}: A spatial compartment model using public mobility data, proposed by Johns Hopkins University.
    \item \textbf{CU-nochange} \citep{a21}: A metapopulation county-level SEIR model, proposed by Columbia University. 
    \item \textbf{COVIDhub-Ensemble} \citep{a22}: An ensemble model containing the best performed models on \emph{The Hub}, proposed by the CDC. 
\end{itemize}

Due to the limit of space, more detailed information of compared models can be found at COVID-19 Forecast Hub official website\footnote{https://zoltardata.com/project/44}. 

\subsection{Implementation Details}
All of our implementations run on a server with an Intel Xeon Platinum 8369B @2.9GHz CPU and a single NVIDIA RTX 2070 Super GPU. To eliminate the data bias originated by demographic differences, we normalize all the data by population factor, encouraging the model to learn the common epidemic patterns sharing across different administrative scales. We set our model to look back 14 days, i.e. we begin training at the beginning of each week with features of last week, producing the inferences for the next week. Moreover, to allow an early stop, we take the latest data of all locations as the validation set. Finally, for the robustness of our proposed model, we utilize different random seeds for training, then averaged the final forecasting result from different seeds. For the N-Beats in temporal convolution module, we set its dimension as 32. As for the graph convolution module, we set the network dimension as 64 with 2 layers, the aggregation type is Max-Pooling, and the node and edge dimensions are 4. The batch size is set as 4 and the learning rate is set as 1e-3. For the graph optimization, we utilize the mini-batch training techniques, the batches are sampled from the whole graph in a certain time period, forming a new sub-graph. In practice, we utilize the random-walk sampling techniques to form the new sub-graphs. Moreover, all the sampled dates are randomly shuffled to avoid information leakage. 

\begin{table}[t]
\caption{A summary of main evaluation results. Best results are highlighted in bold and second best results are highlighted by underlines.}
\label{tab2}
\begin{tabular}{@{}l|c|ccc|ccc@{}}
\toprule
Dataset                              & \multicolumn{1}{l|}{} & \multicolumn{3}{l|}{@500}                         & \multicolumn{3}{l}{@100}                           \\ \midrule
Method                               & Metric                & 1wk            & 2wk            & 3wk             & 1wk            & 2wk             & 3wk             \\ \midrule
\multirow{3}{*}{Microsoft-DeepSTIA}  & MAE                   & 139.0          & 506.7          & 988.4           & 374.5          & 1414.4          & 2628.7          \\
                                     & MAPE                  & 0.425          & 0.455          & 0.607           & 0.392          & 0.461           & 0.604           \\
                                     & RMSE                  & 341.3          & {\ul 980.0}    & {\ul 1808.5}    & 677.2          & {\ul 1992.7}    & 3911.9          \\ \midrule
\multirow{3}{*}{USC-SI\_kJalpha}     & MAE                   & 141.1          & 541.6          & 1045.5          & {\ul 343.1}    & 1440.3          & 2781.6          \\
                                     & MAPE                  & 0.430          & 0.546          & 0.733           & 0.369          & 0.535           & 0.732           \\
                                     & RMSE                  & 366.0          & 1072.7         & 2015.6          & 650.1          & 2081.0          & 3935.7          \\ \midrule
\multirow{3}{*}{UVA-Ensemble}        & MAE                   & 159.7          & 587.6          & 1071.6          & 411.3          & 1570.6          & 2866.9          \\
                                     & MAPE                  & 0.451          & 0.591          & 0.730           & 0.443          & 0.598           & 0.739           \\
                                     & RMSE                  & 418.1          & 1073.6         & 1873.3          & 822.1          & 2179.2          & 3842.6          \\ \midrule
\multirow{3}{*}{Google\_Harvard-CPF} & MAE                   & 178.2          & 685.7          & 1211.8          & 465.6          & 1917.1          & 3346.8          \\
                                     & MAPE                  & 0.456          & 0.627          & 0.738           & 0.368          & 0.623           & 0.767           \\
                                     & RMSE                  & 408.1          & 1352.7         & 2283.2          & 777.9          & 2794.2          & 4718.0          \\ \midrule
\multirow{3}{*}{CEID\_Walk}          & MAE                   & 155.4          & \textbf{485.1} & 969.2           & 409.3          & {\ul 1390.2}    & 2594.6          \\
                                     & MAPE                  & 0.492          & {\ul 0.443}    & {\ul 0.593}     & 0.438          & {\ul 0.447}     & {\ul 0.573}     \\
                                     & RMSE                  & 371.8          & 1021.5         & \textbf{1724.7} & 742.6          & 2016.2          & 3931.2          \\ \midrule
\multirow{3}{*}{IowaStateLW-STEM}    & MAE                   & 321.5          & 694.4          & 1240.5          & 929.0          & 1870.2          & 3316.7          \\
                                     & MAPE                  & 0.729          & 0.550          & 0.644           & 0.654          & 0.554           & 0.669           \\
                                     & RMSE                  & 1091.4         & 1481.9         & 2284.9          & 2312.3         & 3054.7          & 4684.4          \\ \midrule
\multirow{3}{*}{JHUAPL-Bucky}        & MAE                   & 210.5          & 582.1          & 1067.3          & 543.1          & 1501.9          & 2813.2          \\
                                     & MAPE                  & 0.550          & 0.614          & 0.746           & 0.458          & 0.542           & 0.715           \\
                                     & RMSE                  & 530.8          & 1220.1         & 2012.1          & 972.7          & 2154.2          & 3844.2          \\ \midrule
\multirow{3}{*}{CU-nochange}         & MAE                   & 135.1          & 577.3          & 1080.3          & 352.5          & 1585.7          & {\ul 2590.7}    \\
                                     & MAPE                  & {\ul 0.347}    & 0.544          & 0.710           & {\ul 0.285}    & 0.543           & 0.585           \\
                                     & RMSE                  & 327.7          & 1119.2         & 2006.1          & 643.4          & 2304.4          & 4144.3          \\ \midrule
\multirow{3}{*}{COVIDhub-ensemble}   & MAE                   & {\ul 132.2}    & 550.2          & 1040.5          & 343.8          & 1482.2          & 2808.6          \\
                                     & MAPE                  & 0.400          & 0.491          & 0.650           & 0.339          & 0.481           & 0.645           \\
                                     & RMSE                  & {\ul 312.3}    & 1035.9         & 1878.5          & {\ul 599.5}    & 2108.6          & \textbf{3822.6} \\ \midrule
\multirow{3}{*}{MSGNN (Ours)}        & MAE                   & \textbf{121.3} & {\ul 502.2}    & \textbf{959.6}  & \textbf{321.5} & \textbf{1360.4} & \textbf{2588.5} \\
                                     & MAPE                  & \textbf{0.340} & \textbf{0.439} & \textbf{0.584}  & \textbf{0.283} & \textbf{0.432}  & \textbf{0.571}  \\
                                     & RMSE                  & \textbf{302.2} & \textbf{977.8} & 1867.8          & \textbf{594.8} & \textbf{1990.5} & {\ul 3840.9}    \\ \bottomrule
\end{tabular}
\end{table}

\subsection{Main Evaluation Results}
The main evaluation result is reported in Table \ref{tab2}, with best results highlighted in bold and second best highlighted in underline. The evaluation is divided into two datasets, i.e. @500 and @100. For each dataset, we set three forecasting horizons containing one week ahead, two weeks ahead and three weeks ahead, then further calculating their evaluation metrics containing MAE, MAPE and RMSE. All metrics are computed on the US county level, then further averaged to obtain main evaluation results. The same result processing strategies are also applied to other baselines.

As shown in Table \ref{tab2}, our model has superior performance over any other baselines. In most cases, our proposed model produces the best results. For example, when we set the forecasting horizon as one week ahead, our proposed model achieved the lowest MAE, MAPE and RMSE in terms of both @500 and @100 datasets, the best performed baseline forecasting MAE is 132.2 and our proposed model reaches 121.3, bringing in up to 10\% of the performance gain. Moreover, when the forecasting horizon is set as two weeks and more, the proposed model still manages to outperforms the state-of-art methods in terms of MAPE and MAE. Despite the fact that in some cases other baselines achieve better result, our model still manages to achieve the second best score. 

It is also worth noting that for all models, the forecasting accuracy drops when the forecasting horizon becomes longer. For example, the CU-nochange model reaches 0.347 and 135.1 for MAPE and MAE, which is a pretty competitive result in 1-week forecasting scenario. However, when broadening the forecasting horizon, its performance drops sharply to 0.710 and 1080.3. Similarly, the USC-SI\_kJalpha model outperforms most of the baselines in short term forecasting, while producing much poorer results in long term forecasting. Taking all forecasting horizons into consideration, we can observe that the proposed MSGNN still has strong performance. Only for the forecasting in next three weeks, the RMSE metric of MSGNN is relatively larger than other baselines, such as CEID\_Walk and COVIDhub-ensemble. While for all other cases, the proposed model is the best performing model in terms of averaged accuracy. 

\subsection{Per-week Evaluation Results}
Aside from inspecting the main evaluation result, we also notice significant variations in performance across different weeks. Some baseline models may suffer a great performance degradation when facing challenging epidemic patterns, such as sharp increments in confirmed cases, unstable mortality curves, etc. Since these challenging epidemic patterns are widespread in epidemic outbreaks, the sensitivity to these patterns may exert a dramatic negative impact on model robustness. 

To this end, we designed another experiment called per-week evaluation experiment. In this experiment, we collect forecasting error metrics for each model every week from February 2021 to July 2021, and further visualize their distributions to help validate robustness of both proposed model and its competitors.

\begin{figure}
    \begin{minipage}{1\linewidth}
        \centering
        \includegraphics[width=1\linewidth]{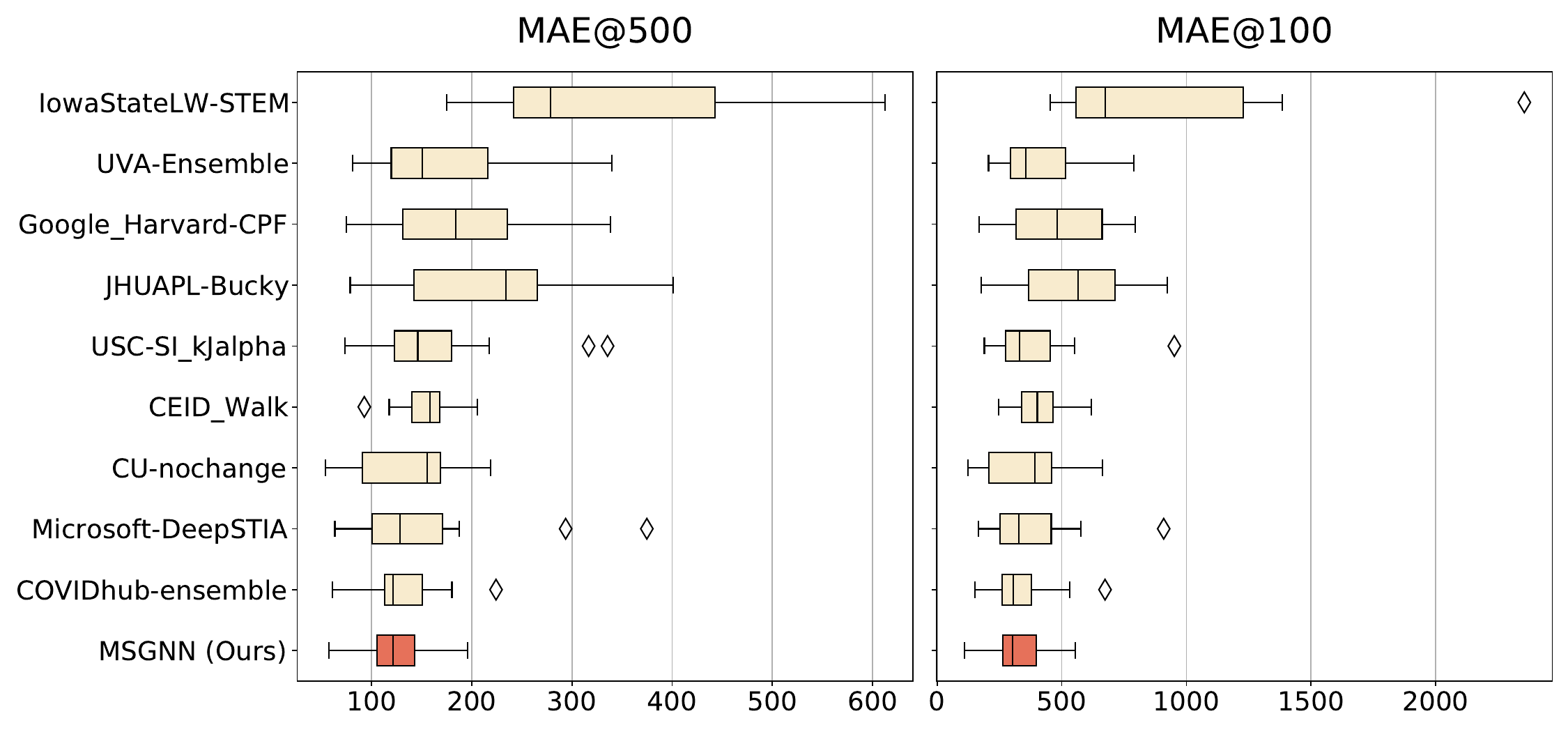}
        \vspace{2mm}
    \end{minipage}
    \begin{minipage}{1\linewidth}
        \centering
        \includegraphics[width=1\linewidth]{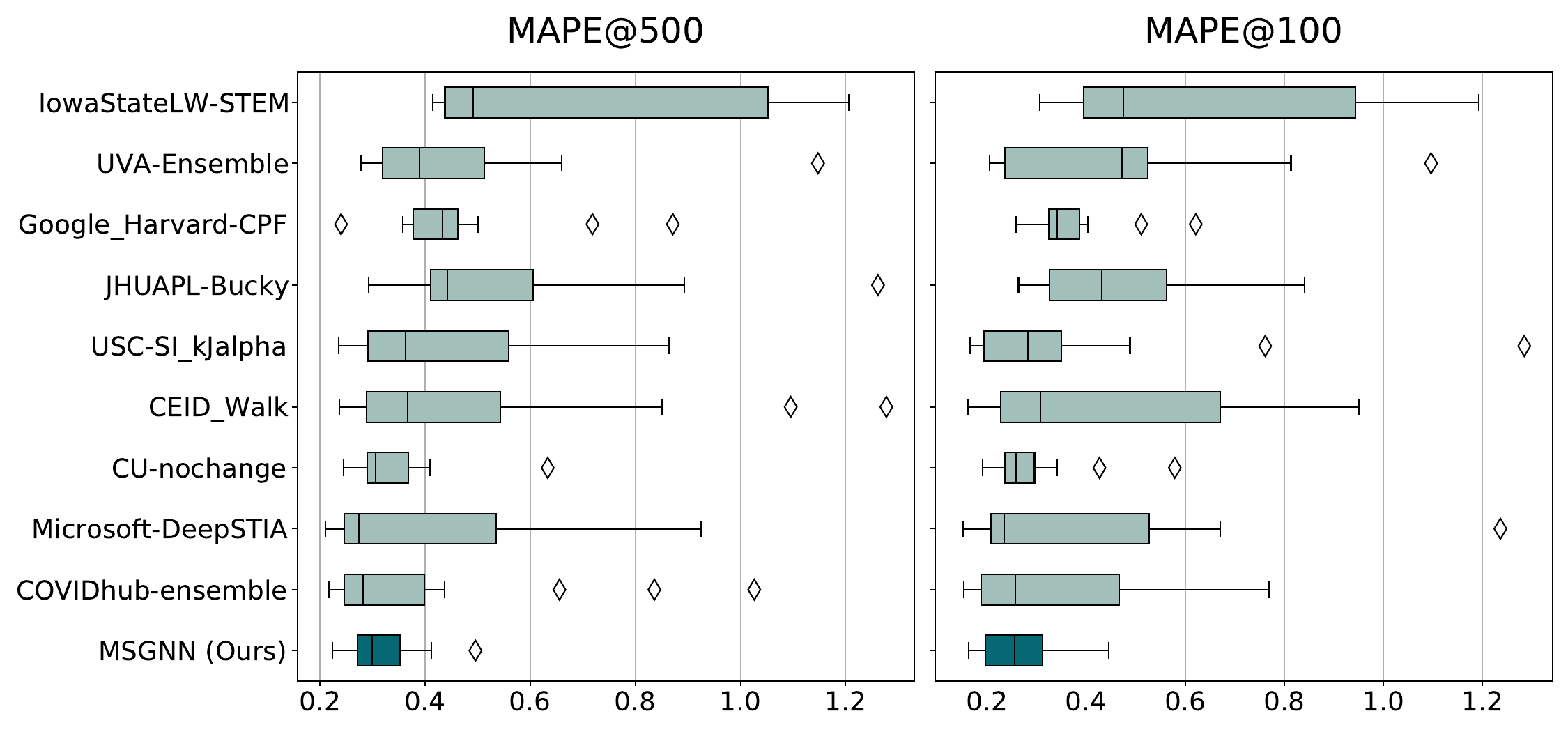}
        \vspace{2mm}
    \end{minipage}
    \begin{minipage}{1\linewidth}
        \centering
        \includegraphics[width=1\linewidth]{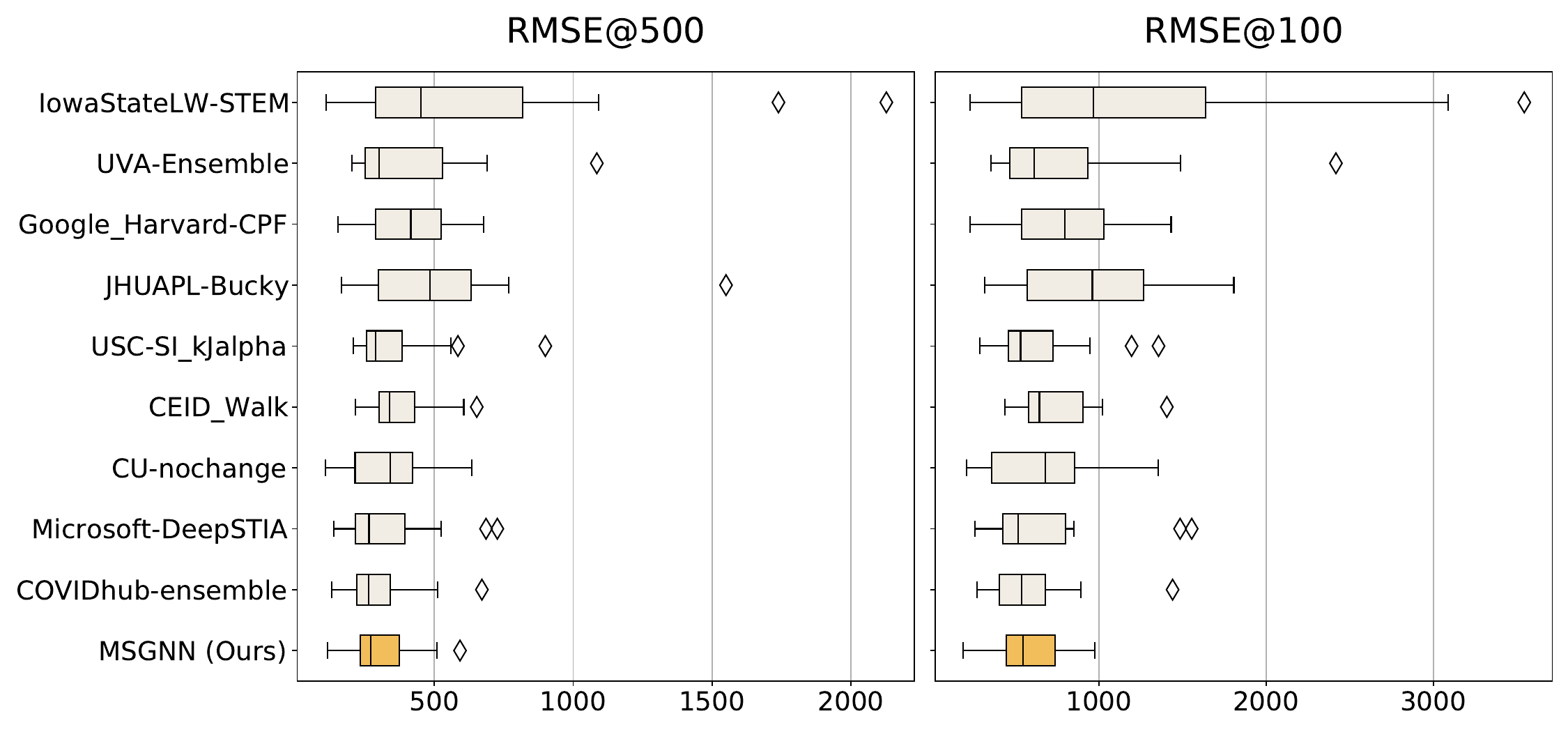}
    \end{minipage}
    \caption{The per-week evaluation results for model robustness experiments, all the results are presented as box plots.}
    \label{fig3}
\end{figure}

As illustrated in Figure \ref{fig3}, the per-week evaluation results of the proposed model is still competitive in terms of MAE, MAPE and RMSE metrics. Our proposed model outperforms other baselines in three perspectives. Firstly, the median value of the proposed model is low. Compared with CEID\_Walk, one of best performed baselines in averaged evaluation experiments, our proposed model reached significantly lower error medians in terms of MAE, MAPE and RMSE. Secondly, the proposed model produces much smaller performance variations. Other competitive models such as Microsoft-DeepSTIA and COVIDhub-ensemble feature larger variations in performance, which means their error distributions are scattered. On the contrary, the proposed model has smaller rectangle size and produces stabler forecasting results compared to other competitors. Finally, MSGNN produces less outliers. The hollow diamonds represent the outliers, whose statistic values are extremely high or extremely low. More outliers will do harm to model's forecasting robustness, leading to lower reliability. Combining the box plots of MAE, MAPE and RMSE, we observe that the total number of outliers produced by MSGNN is the least among all the compared baseline models.  

All the observations prove that when facing challenging epidemic patterns, the proposed model performs more robust than all the baseline models. In summary, the proposed model is the most competitive model producing not only accurate, but also robust epidemic forecasting result. 

\subsection{Ablation Study}
To verify the effectiveness of each component in proposed model, we conduct a comprehensive ablation study, whose result is shown in Table \ref{tab3}. All the experiments are conducted on both @500 and @100 dataset and the results are obtained by averaging the outputs of forecasting 1-week ahead, 2-weeks ahead and 3-weeks ahead. The original version of our proposed model is MSGNN. To further prove the effectiveness of our model, we add another two best performed baselines, i.e. CEID\_Walk and COVIDhub-ensemble for comparison. 

\subsubsection{Effect of the multi-scale modeling} To evaluate the contribution of the multi-scale structure, we create the first variant: 

\textbf{MSGNN-w/o-ms.} MSGNN-w/o-ms is a variant which models epidemic only in county level. To achieve so, we completely remove the additional state level in proposed model, downgrading it to a single-scale spatio-temporal model. Under this circumstance, the model is only accessible to the county level epidemic patterns, thus inevitably leading to worse performance. It can be observed in Table \ref{tab3} that MSGNN-w/o-ms variant produces the worst result among all baselines and variants on both @500 and @100 datasets, which indicates the importance of multi-scale modeling in epidemic forecasting. The reasons why the multi-scale modeling is important could be that without the assistance of information from upper administrative level, the model can only attend to their closest neighbours for information aggregations, ignoring the latent long-range connectivity.

\begin{table}[!tbp]
\centering
\caption{A summary of ablation study results, based on @100 and @500 datasets, averaged by 1-week, 2-weeks, 3-weeks ahead forecasting MAPE.}
\label{tab3}
\begin{tabular}{l|cc|cc}
\hline
Dataset           & \multicolumn{2}{c|}{@100}   & \multicolumn{2}{c}{@500}    \\ \hline
Metric & MAPE & \begin{tabular}[c]{@{}c@{}}Relative\\ Increments\end{tabular} & MAPE & \begin{tabular}[c]{@{}c@{}}Relative\\ Increments\end{tabular} \\ \hline
CEID-Walk         & 0.479          & +5.1\%     & 0.509          & +4.9\%     \\
COVIDhub-ensemble & 0.489          & +6.1\%     & 0.514          & +6\%       \\ \hline
\textbf{MSGNN}     & \textbf{0.428} & \textbf{-} & \textbf{0.454} & \textbf{-} \\
MSGNN-w/o-ms       & 0.509          & +8.1\%     & 0.515          & +6.1\%     \\
MSGNN-w-GCN        & 0.470          & +4.2\%     & 0.491          & +3.7\%     \\
MSGNN-w-GAT        & 0.479          & +5.1\%     & 0.489          & +3.5\%     \\
MSGNN-w/o-fusion & 0.467          & +3.9\%     & 0.475          & +2.1\%     \\ \hline
\end{tabular}
\end{table}

\subsubsection{Effect of graph learning module} 
To understand the role that graph learning module plays in epidemic modeling, we design two variants excluding the graph learning procedures.  

\textbf{MSGNN-w-GCN.} In this variant, we use graph convolution network to replace the graph learning module, i.e. we exclude the dynamic multi-scale spatio-temporal graph and directly use the static geographical distance graph to get the forecasting result, thus forming a naive graph convolution network \citep{a23}. According to the experiment result, the variant performs obviously worse than the original model, which indicates the the dynamic spatio-temporal graph contributes positively to epidemic modeling, enabling the model to dynamically attend to the most relevant regions. Moreover, when applying the naive graph convolution network, the model fails to capture those time-varying dependencies. This is caused by static geographic information serves as the dominant element in spatio-temporal modeling instead of dynamic epidemic interactions. 

\textbf{MSGNN-w-GAT.} Aside from utilizing a static graph, we also develop a variant using graph attention mechanism to construct spatio-temporal graphs. In this variant, we replace the adaptive generation module completely with graph attention network \citep{in19}. As shown in the experiment result, our model outperforms this variant, which proves the effectiveness of self-adaptive adjustment mechanism. Although the graph attention mechanism is capable of capturing part of the dynamic spatial dependencies, however, it fails to take the latent long range spatial dependencies into consideration, leading to limited performance. 

The two variants mentioned above prove the effectiveness of the proposed adaptive graph generation module. With the specifically designed graph learning module, MSGNN can balance both temporal signals and spatial signals. Moreover, the self-adaptive adjustment mechanism assists model distinguishing the most epidemic relevant features, avoiding introducing irrelevant noises. 

\subsubsection{Effect of multi-scale graph convolution module} 

\textbf{MSGNN-w/o-fusion.} In this variant, we remove the multi-scale fusion blocks and directly concatenate local and region features to form the final representations for forecasting. Therefore, this variant is not accessible to scale-specific and scale-shared epidemic patterns. Due to the missing multi-scale epidemic patterns, the performance of this variant is limited compared to the original version. However, it is important to note that although the variant achieves inferior performance over the proposed model, it still outperforms the state-of-art baselines including CEID\_Walk and COVIDhub-ensemble, indicating the multi-scale spatio-temporal graph helps modeling the epidemics better.

\begin{figure}[!t]
    \centering
    \begin{minipage}{0.49\linewidth}
        \centering
        \includegraphics[width=1\linewidth]{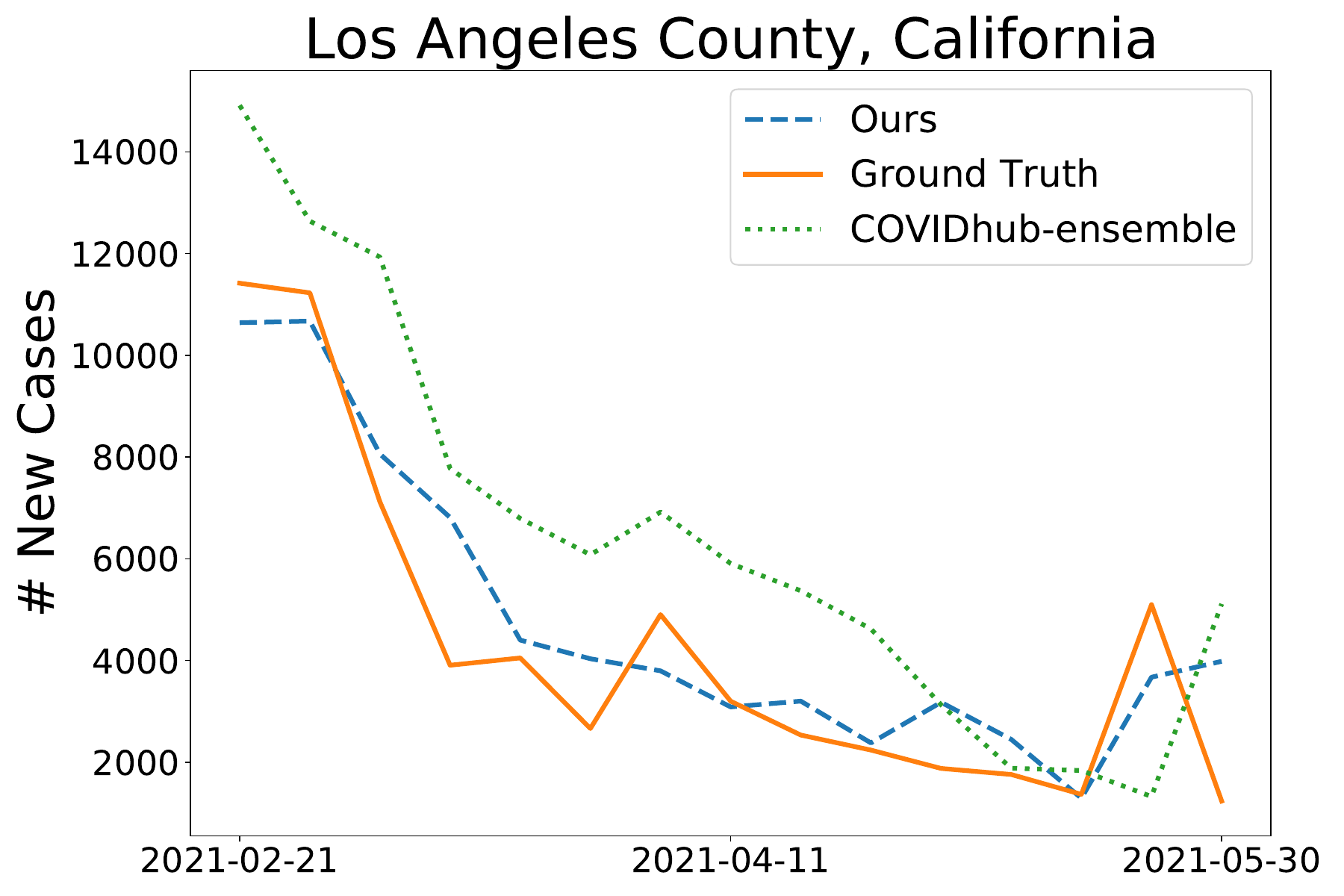}
    \end{minipage}
    \begin{minipage}{0.49\linewidth}
        \centering
        \includegraphics[width=1\linewidth]{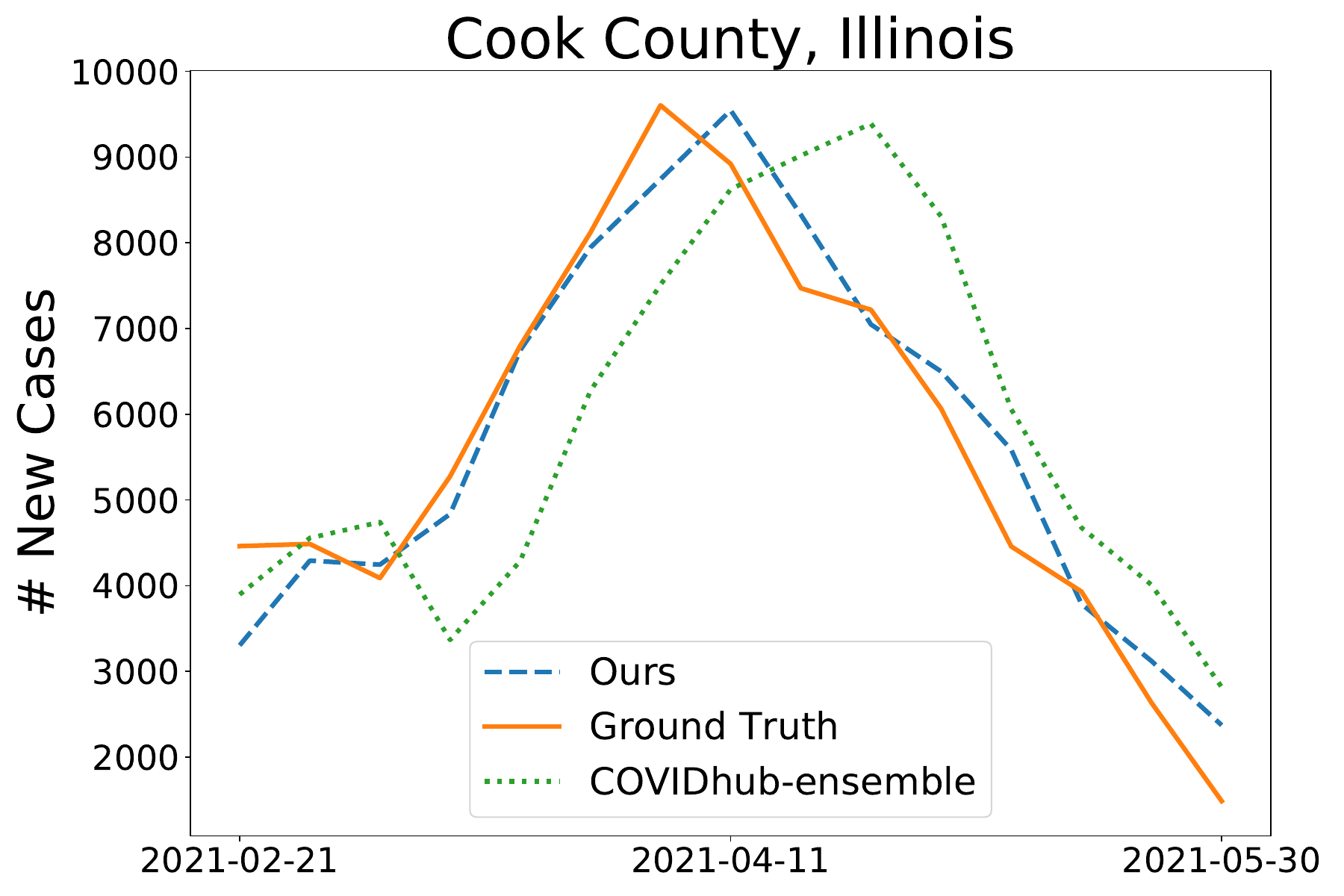}
    \end{minipage}
    \quad
    \centering
    \begin{minipage}{0.49\linewidth}
        \centering
        \includegraphics[width=1\linewidth]{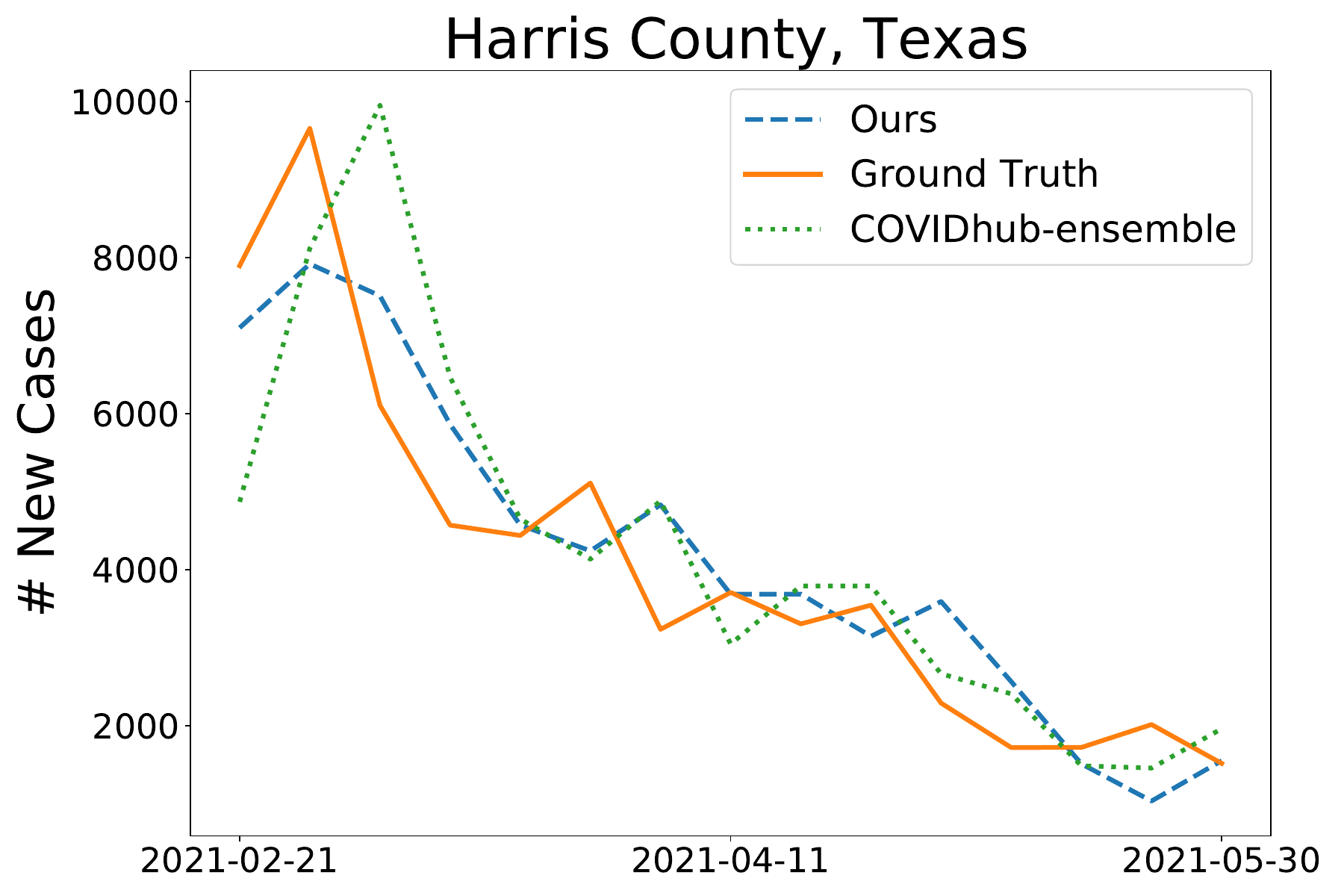}
    \end{minipage}
    \begin{minipage}{0.49\linewidth}
        \centering
        \includegraphics[width=1\linewidth]{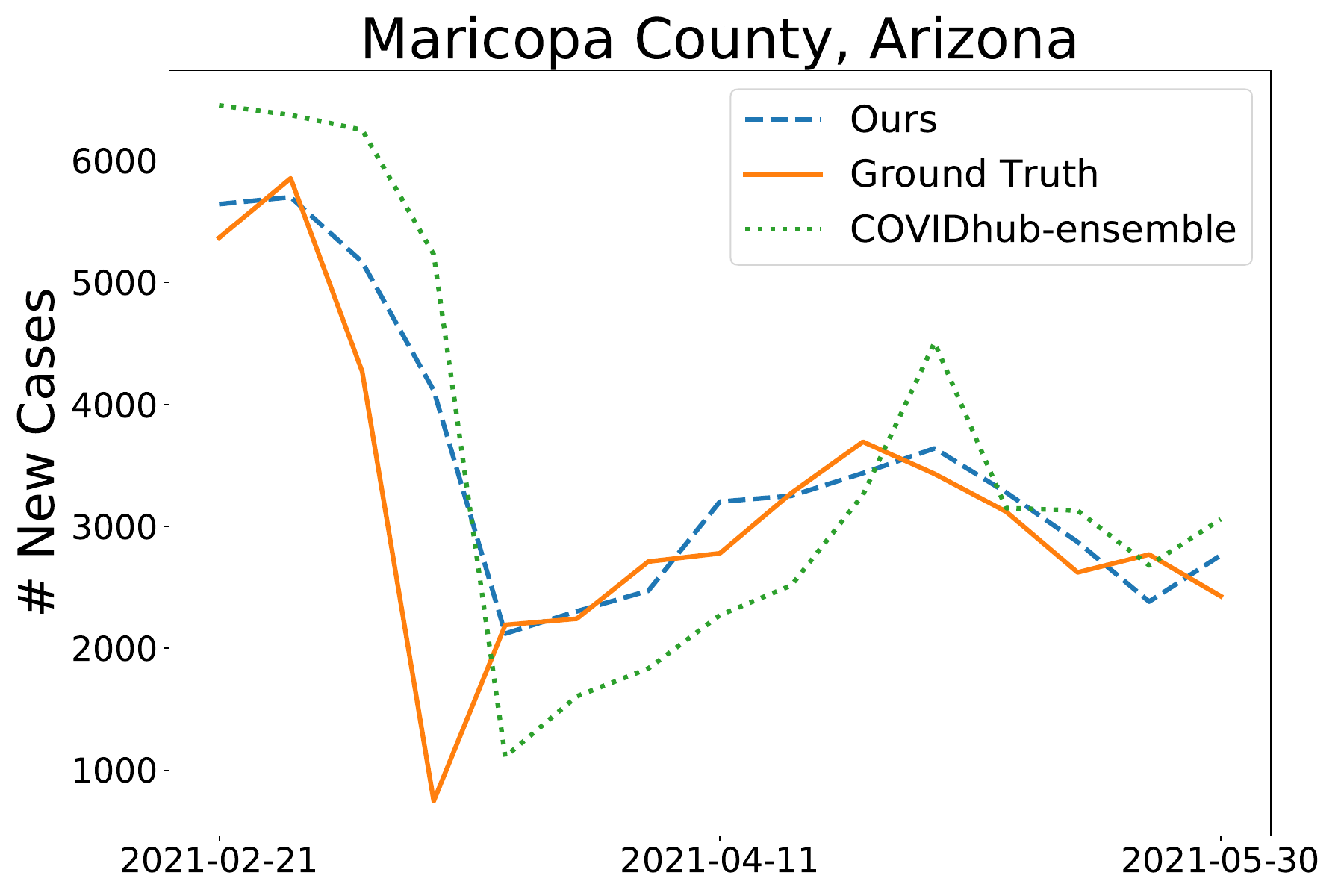}
    \end{minipage}
    \caption{Case studies for four counties with the largest population within a fifteen weeks evaluation period.}
    \label{fig4}
\end{figure}

\subsection{Case Study}
To intuitively understand the evaluation experiments mentioned above, we specifically select four US counties with the largest population to visualize the case study, i.e. Los Angeles County of California, Cook County of Illinois, Harris County of Texas and Maricopa County of Arizona, which are distributed in different parts of the United State. We compare their true incident cases with the forecasting results produced by MSGNN and one of the best performed baseline COVIDhub-ensemble. It is noticeable that COVIDhub-ensemble is an embedding model officially proposed by United State Centers for Disease Control and Prevention (CDC), which comprehensively takes all of the forecasting result from other teams into consideration. 

The result is shown in Figure \ref{fig4}. We can observe that our model still outperforms the COVIDhub-ensemble in all the counties for comparison, producing more accurate and stable forecasting results. In Los Angeles county and Harris county, the trend of ground truth is single and obvious, and the forecasting curve produced by MSGNN fits the ground truth curves better, which means the proposed model achieves better accuracy than COVIDhub-ensemble model. While in the Harris county and Maricopa county, the epidemic trend features more complicated patterns, e.g. larger slope of curves, more slope turning corners, etc. These complex patterns challenge the models' forecasting robustness. It is obvious that the COVIDhub-ensemble model performs poorly with these patterns, where the produced curves are unstable and keep vibrating. Although the proposed MSGNN is also affected by these patterns, it still manages to fit the ground truth in most of the evaluation dates, showing stronger robustness than its competitors. Generally, the overall performance of MSGNN wins more cases than the COVIDhub-ensemble model does.

\section{Discussion: Interpretability Study}
To fully understand the interpretability of our proposed model, in this section, we take a deeper inspect on trans-area and trans-regional epidemic signals produced by graph learning module. Due to the coupled model structure, previous GNN-based methods fail to study model interpretability, which harms forecasting result's credibility. On the contrary, the proposed MSGNN is decoupled into two modules. We can easily study the model's interpretability by inspecting the intermediate output. As the epidemic signals fully control all message passing in the following module, we select the trans-regional and trans-area epidemic signals (i.e. the output of graph learning module), and study the internal relationship between epidemic evolution and these signals.

\begin{figure}[!ht]
    \centering
    \includegraphics[width=1\linewidth]{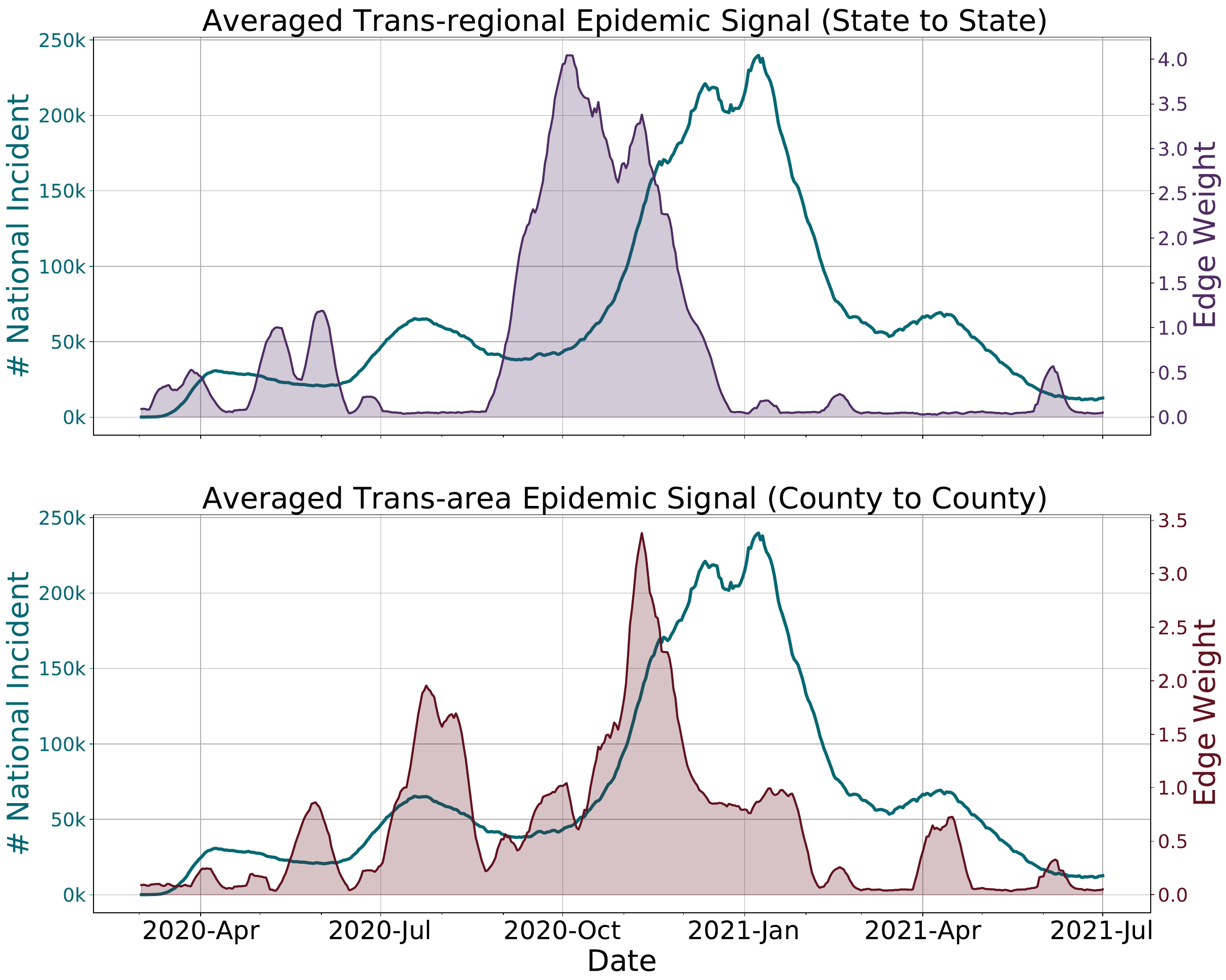}
    \caption{The evolution of spatial dependencies produced by MSGNN (plotted as colored and filled lines) alongside with weekly national incident number (plotted as dark line). }
    \label{fig6}
\end{figure}

We visualize the evolution of epidemic signals alongside national incident numbers. Since our customized graph learning module can dynamically adjust the graph edges based on current epidemic situations, so we directly measures the strength of dependencies through edge weights, where larger edge weights correspond to more active signals. We record all the edge weights produced by a fully trained MSGNN for each forecasting date and present the results in two line charts. as shown in Figure \ref{fig6}. 

Firstly, for long-range connectivity, we can observe that the signals tend to become activated before next wave of sharply increasing new cases. For example, the first peak of activated signals appears in March 2020, which is followed by a rapidly increasing wave of incident cases in April 2020. This kind of pattern is not unique and recur in May 2020, June 2020, October 2020 and February 2021. By aligning the evolution of long-range connectivity with national incident curve, we suggest this interesting pattern reflects the potential transmission process between distant regions. The more active trans-regional epidemic signals indicate the more benefits of spatial signals in epidemic forecasting, which also implies more distant transmissions in the physical world. 

Secondly, it is obvious that the active period of short-range dependency almost covers the peak of long-range connectivity. Especially since January 2021, the trans-area signals have much more distinct peak values than long-range connectivity. We conjectured that as the pandemic is gradually controlled in 2021, the large scale nation-wise transmissions were successfully suppressed, leading to inactivated long-range connectivity signals. However, the local epidemic transmissions still exist and forms active short-range epidemic signals. Hence, we suggest the short-range dependency indicates the local transmission process among neighbouring areas. 

In summary, the proposed MSGNN has some interpretable capacities as it explicitly shows how epidemic transmit at different spatial scales. On one hand, the learned epidemic signals help modeling epidemic better and ensure the credibility of forecasting result. On the other hand, it also installs confidence in end-users such as policy makers and citizens. 

\section{Conclusions}
In this paper, we propose Multi-scale Spatio-temporal Graph Neural Network (MSGNN) for epidemic forecasting. Specifically, we design a novel multi-scale epidemic modeling method. On one hand, we devise a graph learning module to directly captures long-range connectivity and integrate them into a multi-scale spatio-temporal graph. On the other hand, we customized a multi-scale graph convolution module, which adopts a novel information aggregation scheme. The scheme allows us to distinguish both scale-specific and scale-shared epidemic patterns, which gives rise to multi-scale modeling. Extensive experiments are conducted and the results demonstrate the capacity of MSGNN from multiple dimensions, including the forecasting accuracy, robustness and forecasting interpretability.

\backmatter

\section*{Declarations}

\bmhead{Funding}
This work was supported by National Key Research and Development Project (No.2020AAA0106200), the National Nature Science Foundation of China under Grants (No.61936005, 61872424), the Natural Science Foundation of Jiangsu Province (Grants No.BK20200037 and BK20210595).

\bmhead{Conflict of interests}
The authors have no conflict of interest to declare that are relevant to the content of this article. 

\bibliography{sn-bibliography}

\end{document}